\documentclass[a4paper,fleqn]{cas-dc}

\usepackage[numbers]{natbib}
\usepackage{graphicx}
\usepackage{float}
\usepackage{subfigure}
\usepackage{soul}
\usepackage{amsmath}
\usepackage{amssymb}
\usepackage{xcolor}
\usepackage{hyperref}
\soulregister{\cite}7 
\soulregister{\citep}7 
\soulregister{\citet}7 
\soulregister{\ref}7 
\soulregister{\pageref}7 

\def\tsc#1{\csdef{#1}{\textsc{\lowercase{#1}}\xspace}}
\tsc{WGM}
\tsc{QE}
\tsc{EP}
\tsc{PMS}
\tsc{BEC}
\tsc{DE}

\makeatletter
\def\ps@first{%
   \let\@oddhead\@empty
   \let\@evenhead\@empty
   \def\@oddfoot{}
   \let\@evenfoot\@oddfoot
}
\begin{document}
\let\WriteBookmarks\relax
\def\floatpagepagefraction{1}
\def\textpagefraction{.001}
\bibliographystyle{plainnat}
\renewcommand{\thefootnote}{\relax}


\shorttitle{A New Prompting Multi-modal Model Paradigm}

\shortauthors{Z. Wang et~al.}

\title [mode = title]{PM$^{2}$: A New Prompting Multi-modal Model Paradigm for Few-shot Medical Image Classification}

%
\author[1,2,3]%
{Zhenwei Wang}
\cormark[1]

\affiliation[1]{organization={Key Laboratory of Social Computing and Cognitive Intelligence(Ministry of Education)},
    addressline={Dalian University of Technology}, 
    city={Dalian},
    postcode={116024}, 
    country={China}}

\affiliation[2]{organization={School of Computer Science and Technology},
    addressline={Dalian University of Technology}, 
    city={Dalian},
    postcode={116024}, 
    country={China}}

\affiliation[3]{organization={School of Computer Science and Engineering},
    addressline={Dalian Minzu University}, 
    city={Dalian},
    postcode={116600}, 
    country={China}}

\affiliation[4]{organization={School of Information and Communication Engineering},
    addressline={Dalian University of Technology}, 
    city={Dalian},
    postcode={116024}, 
    country={China}}


\author[4]%
{Qiule Sun}
\cormark[1]

\author[3]{Bingbing Zhang}
\author[1,2]{Pengfei Wang}
\author[3]{Jianxin Zhang}[orcid=0000-0001-6076-5433]\cormark[2]
\ead{jxzhang@dlnu.edu.cn}

\author[1,2]{Qiang Zhang}[orcid=0000-0003-3776-9799]\cormark[2]
\ead{zhangq@dlut.edu.cn}

\cortext[cor1]{The first two authors contributed equally to the work.}
\cortext[cor2]{Corresponding author}

\begin{abstract}
Few-shot learning has been successfully applied to medical image classification as only very few medical examples are available for training. Due to the challenging problem of limited number of annotated medical images, image representations should not be solely derived from a single image modality which is insufficient for characterizing concept classes. In this paper, we propose a new prompting multi-modal model paradigm on medical image classification based on multi-modal foundation models, called PM$^{2}$. Besides image modality, PM$^{2}$ introduces another supplementary text input, known as prompt, to further describe corresponding image or concept classes and facilitate few-shot learning across diverse modalities. To better explore the potential of prompt engineering, we empirically investigate five distinct prompt schemes under the new paradigm. Furthermore, linear probing in multi-modal models acts as a linear classification head taking as input only class token, which ignores completely merits of rich statistics inherent in high-level visual tokens. Thus, we alternatively perform a linear classification on feature distribution of visual tokens and class token simultaneously. To effectively mine such rich statistics, a global covariance pooling with efficient matrix power normalization is used to aggregate visual tokens. Then we study and combine two classification heads. One is shared for class token of image from vision encoder and prompt representation encoded by text encoder. The other is to classification on feature distribution of visual tokens from vision encoder. Extensive experiments on three medical datasets show that our PM$^{2}$ significantly outperforms counterparts regardless of prompt schemes and achieves state-of-the-art performance.
\end{abstract}


\begin{keywords}few-shot learning\sep multi-modal modeling\sep prompt learning\sep covariance pooling\sep medical image classification
\end{keywords}

\maketitle

\section{Introduction}

\footnotetext{This work has been submitted to the IEEE for possible publication. copyright may be transferred without notice, after which this version may no longer be accessible.}

Medical imaging constitutes a pivotal tool in medical diagnostics, encompassing diverse modalities such as X-rays, CT scans, MRI, and ultrasound~\cite{panayides2020ai}. Medical imaging confers numerous benefits to disease diagnosis, including non-invasive diagnostics, early disease detection, heightened precision and accuracy, as well as assisting in treatment planning and disease progression monitoring. In the medical imaging analysis, clinicians face the challenge of subjectivity, as it heavily depends on the individual experience and judgment of the physician. This inherent subjectivity can result in divergent diagnoses among different experts when interpreting the same image. Utilizing computer-aided diagnosis offers a promising approach to address the aforementioned challenges~\cite{yadav2019deep}. Employing deep learning and allied technologies in the domain of medical image diagnosis present a prospect for elevating diagnostic precision while expeditiously handling substantial volumes of medical image data. Concurrently, it addresses inherent issues of subjectivity and fatigue often encountered by human practitioners.

The past decade has witnessed a robust surge in the advancement of deep learning, with deep learning techniques attaining substantial success in domains such as computer vision and natural language processing (NLP)~\cite{fink2020potential}. Notably, certain researchers have utilized deep learning methodologies to tackle a range of challenges within the medical sector, with a particular emphasis on medical image classification, which stands as one of the most extensively explored domains. Through the implementation of a series of best practices, such as leveraging insights from successful natural image classification to craft a model tailored for medical images, fine-tuning pre-trained models with medical image datasets, and using data augmentation to expand data, these approaches have yielded notable outcomes~\cite{kora2022transfer}. They have even, to some extent, attained classification accuracies comparable to those of expert physicians. Nonetheless, it is crucial to acknowledge that these approaches collectively confront an incontrovertible reality: their dependence on a substantial corpus of meticulously annotated examples for training, a requisition that presents impracticalities within the realm of medicine~\cite{wang2021deep}. Acquiring a vast repository of medical images and meticulously annotating them is a prohibitively expensive endeavor, necessitating specialized equipment for image acquisition and the manual labeling conducted by practitioners knowledge both professional acumen and extensive experiences~\cite{cheplygina2019not,litjens2017survey}. Consequently, the majority of deep learning-based methodologies face formidable barriers when applied to practical medical tasks, due to the aforementioned challenges.

Few-shot learning~\cite{finn2017model,bateni2020improved, snell2017prototypical,qi2018low} represents an advanced iteration within the domain of deep learning, engineered to distill the essence of entire data categories from sparse datasets. This evolution effectively circumvents the inherent challenges that conventional machine learning approaches grapple with when confronted by constraints of scant training data. The advantages of few-shot learning make it quickly become a popular and promising topic in the research field of medical image classification~\cite{nayem2023few,chen2021momentum,dai2023pfemed,jiang2022multi} as each class only has a very limited number of examples. Chen et al.~\cite{chen2021momentum} employed contrastive learning to train an encoder capable of capturing highly expressive feature representations on an extensive and publicly accessible lung dataset. Subsequently, they harnessed a prototypical network for the classification of COVID-19 CT images. Jiang et al.~\cite{jiang2022multi} devised real-time data augmentation and dynamic gaussian disturbance soft label (GDSL) scheme as effective generalization strategies of few-shot classification tasks. Dai et al.~\cite{dai2023pfemed} introduce a novel prior-guided variational autoencoder (VAE) module to enhance the robustness of the target feature, which is the concatenation of the general and specific features. These few-shot learning methods are first pre-trained on a large meta-training dataset, which enables the model to learn how to extract general knowledge from multiple tasks. They are then evaluated on different training (support) sets and test (query) sets to assess their adaptability and generalization capabilities for new tasks. However, the meta-learning phase typically requires substantial computational resources and time. CLIP (Contrastive Language-Image Pre-Training)~\cite{radford2021learning} is a multi-modal model trained through contrastive learning, enabling zero-shot recognition without the need for class-specific examples during the learning process. In our study, we draw inspiration from recent advances~\cite{radford2021learning,zhang2021tip,zhou2022learning} in CLIP to transition to a novel few-shot evaluation protocol where we replace the meta-training stage with a pretrained CLIP model. 

\begin{figure*}[t]
    \centering
    \includegraphics[scale=0.62]{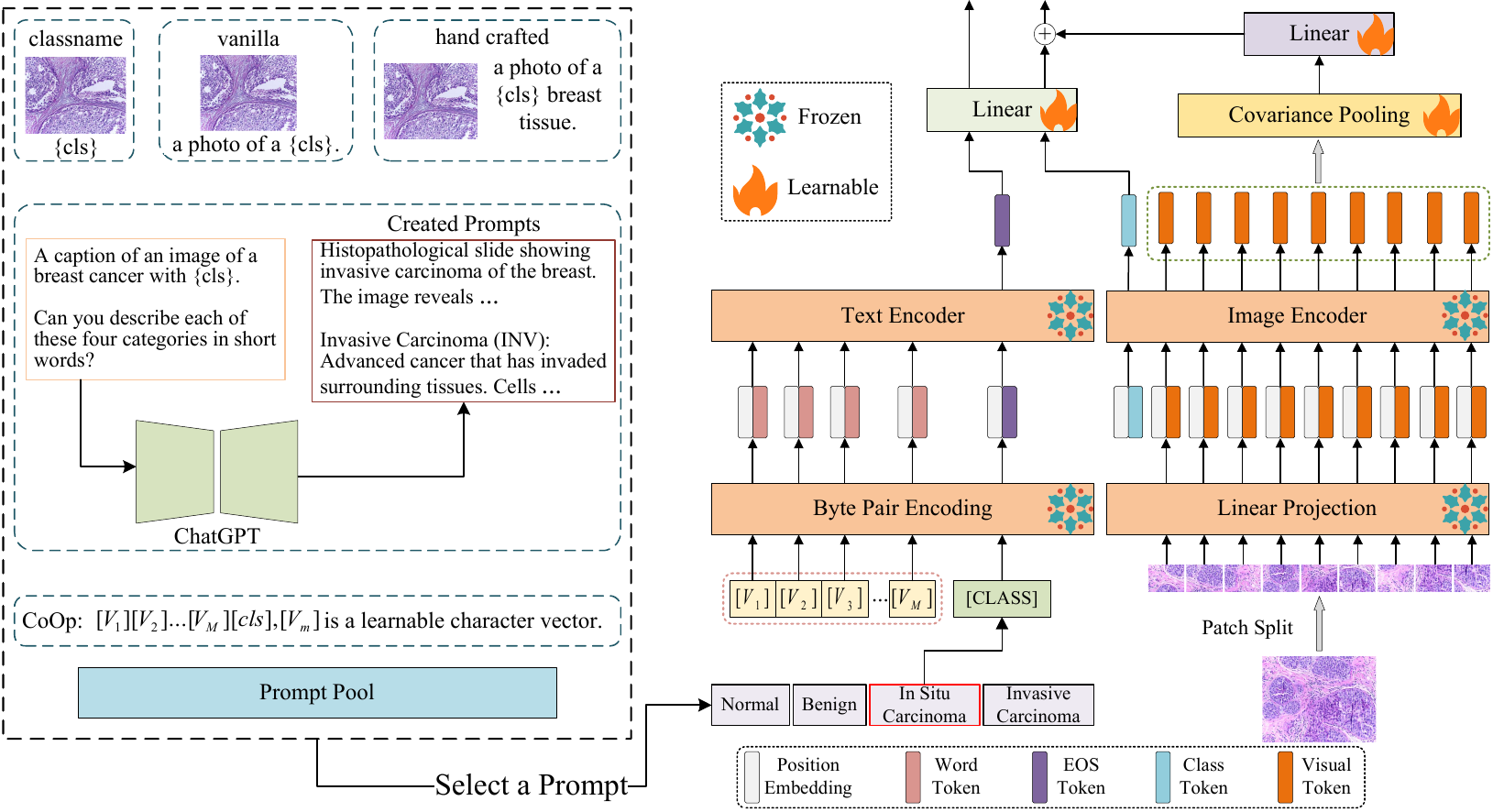}
    \caption{Schematic diagram of the PM$^2$ structure. First, a prompting method is selected from the Prompt Pool to generate prompts for each class, using the CoOp method as example in the diagram. Then, visual and textual features are extracted using their respective encoders, with ViT as example of a visual encoder. Subsequently, a shared classifier predicts classes based on text features ([EOS] token) and a visual class token. For the visual part's class prediction, covariance pooling of visual tokens is further calculated for second-order statistical modeling. The second-order prediction is then added to the first-order prediction, represented by a single class token, to form the final prediction for the visual component. The text encoder and image encoder are frozen during training for efficiency.}
    \label{fig1}
\end{figure*}

Motivated by recent advances in multi-modal learning research, we propose a new prompting multi-modal model paradigm, PM$^{2}$, as shown in Fig.~\ref{fig1}. In our pipeline, prompt is introduced non-trivially to provide a complementary alternative to sufficiently describe images or their corresponding concept classes. Regarding the design of prompts, our prompt pool includes five prompt schemes, as it is challenging to describe the content or categories of medical images using a vanilla prompt template like "a photo of a \{cls\}" in CLIP. Beyond the vanilla prompt, the other four prompt templates are: classname, hand-crafted~\cite{zhang2021tip}, using GPT (Generative Pre-trained Transformer)~\cite{brown2020language} to generate descriptions for each class~\cite{zhang2023prompt}, and employing CoOp~\cite{zhou2022learning} to learn text prompts for each class. Such text prompt (text modality) can better help few-shot learning in conjunction with images. Our method is constructed on the foundation of CLIP, benefiting from its successful alignment of text and image modalities. Any general-propose, dual-encoder multi-modal models pre-trained on vision-language datasets can also be an alternative one, since we focus on the proposed paradigm in few-shot learning under medical image classification scenario rather than the specific instance backbone. For efficiency, the employed CLIP model is frozen in our task and only a linear classifier (i.e., linear probing~\cite{he2020momentum,radford2021learning}) is turned. Compared to full fine-tuning, linear probing usually suffers from inferior performance. But full fine-tuning usually needs a large number of labeled medical samples which are prohibitively unavailable in practice, and suffers from too much computation burden as well. To show efficiency and effectiveness trade-off in our paradigm, we adopt linear probing but boost its ability by taking as input feature distribution of input features.

We observe that linear probing learns a linear classifier on input features which are outputted from convolutional neural networks (CNN) or vision transformer (ViT) architectures. The input features can be convolution ones, class token, or average pooling (mean) of visual tokens. Regardless of which architecture the input features come from, from the statistical perspective, they merely capture the first-order statistics connected to linear probing. To this end, for effectiveness, going beyond the first-order feature statistics, we mine rich (second-order) statistical information inherent in input features for generating more powerful representations fed into linear probing. In addition, previous works~\cite{li2018towards,xie2022joint,xie2021sot}, incorporating second-order pooling in their networks, become instrumental in comprehending data structure and feature interrelationships. In various instances, it has demonstrated superior performance to first-order methods in tasks like image classification and has been substantiated as efficacious in the domain of medical image classification~\cite{wang2021second,zou2022breast}. Drawing inspiration from these precedents, we consider modeling such powerful feature distribution into our paradigm. Specifically, we model probability density of features (visual tokens) from vision encoder by their characteristic function, and approximate it by using the statistical moment. For computational efficiency, we only consider the second-order moment of features, since the class token represents the first-order statistical moment and is combined with visual tokens to form our visual classification head. After that, matrix power normalization on the second-order moment is adopted to boost its modeling ability. So for, linear probing takes as input feature distribution as stronger representations and makes the prediction more precise.

Our contributions can be summarized as the following:

(1) We propose a new prompting multi-modal model paradigm, PM$^2$, on few-shot medical image classification. To the best of our knowledge, this is the first empirical study to date that considers text prompts as supplementary training samples or modality to describe images or concept classes.

(2) We do an in-depth investigation of five text prompts to comprehensively assess their impact on medical image classification task under the few-shot setting. All of the involved text prompts can help improve the use of only a single image modality, spanning and benefiting architectures of CNN and transformer as well as different datasets.

(3) We introduce a novel visual classification head on the top of vision encoder, which performs a linear classifier on stronger image representations in visual tokens characterized by feature distribution. The visual classification head also combines the the first-order statistics of class token and the second-order statistics of visual tokens simultaneously for classification.

(4) We perform thorough ablation study to validate our PM$^2$. Extensive experiments on three medical image datasets demonstrate that our approach outperforms other counterparts and achieves state-of-the-art performance.

\section{Related Work}
\subsection{Medical Image Classification}
Initially, medical image classification relied on hand-crafted features, but with the advent of CNNs such as AlexNet~\cite{litjens2017survey}, it shifted towards deep learning and automatic feature extraction, significantly enhancing accuracy~\cite{shen2017deep,zhou2023medical,hasan2023fp,qureshi2023medical}. However, CNNs tend to focus on local details, often missing the global context. In contrast, transformers, with their multi-head self-attention mechanism, excel at capturing global information, demonstrating superior performance in medical image analysis~\cite{zhou2023self,komorowski2023towards,li2023new,shamshad2023transformers}. Yet, each approach exhibits limitations in balancing local and global features. Recent methods combine CNNs and transformers, aiming to achieve an optimal balance~\cite{yuan2023effective,manzari2023medvit,ding2023ftranscnn}. Many medical image classification methods utilize supervised learning, but the challenges in obtaining and annotating large-scale medical images limit their development. Consequently, learning effectively from a few-shot becomes essential. Few-shot learning, which operates on a very limited dataset, addresses the need for large-scale data in traditional training. Applied in medical image classification~\cite{prabhu2019few,finn2017model,bateni2020improved,snell2017prototypical,qi2018low,dai2023pfemed}, this approach offers a solution to data scarcity, aiming to develop robust models with minimal datasets.

Although many few-shot classification techniques for medical images utilize the full dataset, this approach does not fully align with the concept of few-shot learning. Our study limited the sample size to a maximum of 16 instances per category, instead of using a larger number of samples for training. This strategy addresses not only the challenge of sample collection but also helps mitigate class imbalance by ensuring uniform sampling across categories.

\subsection{Vision-Language Models}
Building on the success of text pre-training with large-scale web data, researchers are now shifting towards multi-modal contrastive pre-training using massive, noisy image-text pairs. This approach aims for better alignment of image and text representations. Key models such as CLIP~\cite{radford2021learning} and ALIGN~\cite{jia2021scaling} utilize 400 million and 1 billion pairs, respectively. SimVLM~\cite{wang2021simvlm} focuses on multi-modal sequence generation, while FLAVA~\cite{singh2022flava} combines contrastive and generative pre-training for both single and multi-modal tasks. UniCL~\cite{yang2022unified} and CoCa~\cite{yu2022coca} utilize both annotated and web-crawled data, and OmniVL~\cite{wang2022omnivl} integrates image and video language tasks. In visual-language (V-L) model fine-tuning, particularly for few-shot image recognition, research demonstrates improved performance with specialized techniques~\cite{gao2021clip,zhang2021tip}. CLIP-Adapter~\cite{gao2021clip} introduces a lightweight residual feature adapter for fine-tuning, while Tip-Adapter~\cite{zhang2021tip} employs a key-value cache model for adapter weights. Prompt-based learning, such as CoOp~\cite{zhou2022learning}, focuses on optimizing continuous prompt vectors, and CoCoOp~\cite{zhou2022conditional} addresses generalization issues by conditioning prompts on specific images. Another approach~\cite{lu2022prompt} optimizes diverse prompt sets by learning prompt distributions.

When properly applied, these advanced techniques outperform basic CLIP features, significantly enhancing performance. CLIP establishes a new standard for the semantic joint representation of pre-trained images, and fine-tuning adapts it to various tasks. However, the application of these techniques in medical image analysis remains limited. Inspired by these advancements, our study introduces a multi-modal approach specifically tailored for medical images. We experiment with various text prompt methods and incorporate high-order feature modeling in the visual domain, yielding enhanced and robust visual representations.
\subsection{Prompt Learning}
The concept of the prompt was first extensively studied in the field of natural language processing~\cite{liu2023pre}, involving the use of task-specific prompts to augment large pre-trained models for adapting to new tasks. For example, GPT~\cite{radford2018improving,radford2019language,brown2020language} excels in various transfer learning tasks, including zero-shot and few-shot scenarios, by leveraging carefully selected prompts. Prompt learning enables predictions based solely on prompts without updating model parameters, thus facilitating the easier application of large pre-trained models in real-world tasks. Recently, prompt in visual language modeling has also received extensive study. Common forms include vanilla prompts~\cite{radford2021learning} and hand-crafted prompts~\cite{zhang2021tip}. CoOp~\cite{zhou2022learning} and CoCoOp~\cite{zhou2022conditional} outperform hand-crafted prompts by learning continuous contextual vectors, adapting pre-trained V-L models to downstream tasks. CaFo~\cite{zhang2023prompt} leverages GPT-3~\cite{brown2020language} to generate textual inputs, thereby prompting V-L models with rich downstream language semantics.

Compared to research in general domains, studies on prompt learning in medical image classification are less common. Drawing on the success of prompt learning in natural language processing and visual language modeling, we have developed various prompting schemes for this task. This includes manually designed classname, vanilla, and hand-crafted methods. Additionally, we explore a method of generating answers to data category-related questions using GPT and employing these detailed answers as prompts. To fully exploit the potential of prompt learning, further effective generalization of medical images has been achieved through learnable contextual vectors during the fine-tuning process.

\subsection{Second-order Pooling}
The global covariance (second-order) pooling (GCP) has been successfully applied to various tasks and has shown good performance compared to global average (first-order) pooling. B-CNN~\cite{lin2015bilinear} is one of the earliest end-to-end covariance pooling networks. It models local feature interactions by computing the outer product of two convolutional networks. 
MPN-COV~\cite{li2017second} proves that power normalization of the covariance matrix can effectively exploit the geometry of the matrix.
Subsequently, Li et al.~\cite{li2018towards} propose an iterative matrix square root normalization on GCP to accelerate the training of MPN-COV. 
TCPNet~\cite{gao2021temporal} and DeepBDC~\cite{xie2022joint} have shown its potential on challenging video recognition and few-shot classification by introducing temporal-attentive covariance pooling and deep Brownian distance covariance respectively. Recently, to balance representation decorrelation and information preservation, rather than post-normalization on GCP, DropCov~\cite{wang2022dropcov} is proposed for pre-normalization on it via a noval adaptive channel dropout and show its effectiveness on both deep CNN and ViT architectures.
From an optimization perspective, Wang et al.~\cite{wang2023towards} explain and analysis the underlying reasons why global covariance pooling is an effective alternative across networks and tasks.

There are some ways to enhance the representation of visual semantic information in medical images by using high-order feature modeling~\cite{wang2021second,zou2022breast,li2020breast} to obtain higher classification accuracy. Nonetheless, it is noteworthy that these approaches predominantly function within a single modality and are contingent upon the utilization of the complete training dataset. 
In contrast, our methodology focuses on two modalities of image and text for individual prediction of them in the few-shot settings. Moreover, the proposed visual classification head employs both first-order and second-order information of image features for prediction.


\begin{table*}[H]
\centering
\caption{Details of classname, vanilla, hand-crafted, GPT and CoOp prompt methods.}\label{tab1}
\begin{tabular}{ll} \hline
Prompt Method & Details \\ \hline
\multicolumn{1}{c}
{classname}   &  Normal, Benign, In Situ Carcinoma, Invasive Carcinoma \\ \hline
\multicolumn{1}{c}
{vanilla}  & a photo of a \{cls\}.  \\ \hline
{hand-crafted} & a photo of a \{cls\} breast tissue. \\ \hline
GPT Prompt 0  & \begin{tabular}[c]{@{}l@{}}Q1: Do you know the BACH breast cancer dataset?\\ 
A1: Yes, I'm familiar with the BACH (Breast Cancer Histology) dataset. The BACH dataset is a collection of \\ histopathology images of breast cancer biopsies, designed ...\\ 
Q2: Can you describe each of these four categories in short words?\\ 
A2: Benign (B): Non-cancerous tissue. It typically shows normal cellular structure and lacks the aggressive \\ characteristics seen in cancerous tissue.\\ 
In Situ Carcinoma (CIS): Early-stage cancer that hasn't spread beyond its original location. Cells appear abnormal \\ and are confined to their site of origin.
\\ Invasive Carcinoma (INV): Advanced cancer that has invaded surrounding tissues. Cells display aggressive behavior, \\ potentially spreading to other parts of the body.
\\ Normal (N): Healthy, non-cancerous breast tissue that serves as a baseline for comparison with cancerous tissues.\end{tabular}\\ \hline
GPT Prompt 1   & \begin{tabular}[c]{@{}l@{}}A caption of an image of a breast cancer with invasive carcinoma.\\ "Histopathological slide showing invasive carcinoma of the breast. The image reveals irregularly shaped cancerous \\ cells infiltrating surrounding breast tissue, indicative of aggressive invasive carcinoma. Staining highlights \\ the abnormal cellular growth, emphasizing the urgent need for diagnosis and treatment."\\
A caption of an image of a breast cancer with in situ carcinoma.\\ 
"Microscopic view of breast tissue displaying in situ carcinoma. The image depicts cancerous cells confined within \\ the milk ducts or lobules, without invading the surrounding tissue. This non-invasive form of carcinoma highlights \\ the importance of early detection and intervention for effective treatment."\\ 
A caption of an image of a breast cancer with Benign.\\ 
"Microscopic examination revealing benign breast tissue. The image showcases normal and non-cancerous cellular \\ patterns, underscoring the absence of malignant growth. Accurate differentiation between benign and \\ malignant conditions aids in proper medical management and provides reassurance."\\ 
A caption of an image of a breast tissue with Normal.\\ 
"Microscopic snapshot of healthy breast tissue. The image showcases well-organized and non-aberrant cellular \\ structures, representing a normal and functioning breast tissue composition. Understanding the baseline \\ appearance of normal tissue is crucial for identifying any deviations that may warrant further medical assessment."\end{tabular}
\\ \hline
CoOp & $[V]_{1}[V]_{2}...[V]_{M}[cls]$, $[V]_{m}$ is a learnable character vector.\\ \hline
\end{tabular}
\end{table*}

\section{Method}
In this section, we will provide a detailed introduction to the method presented in this paper. First, we review CLIP in Sec.~\ref{s2.0}. Second, in Sec.~\ref{s2.1}, we will outline the overall framework of our approach. Then, the text prompt of this paper is described in Sec.~\ref{s2.2}, and finally the visual classification head is introduced in Sec.~\ref{s2.3}.

\subsection{Revisiting CLIP}
\label{s2.0}
Our approach is based on the pre-trained V-L model CLIP~\cite{radford2021learning}, which includes both text and visual encoders. Thus, a brief introduction to CLIP, standing for Contrastive Language-Image Pretraining, is essential before exploring our method. CLIP processes an image $\mathbf{I}\in \mathbb{R}^{H\times W\times 3}$ along with its associated textual description.

\textbf{Encoding Image:} The image encoder of CLIP can utilize architectures of ResNet~\cite{he2016deep} or ViT~\cite{dosovitskiy2020image}. This encoder functions to transform visual data into feature vectors. Let's represent the visual encoder as $\mathcal{V}$. When ViT is used for $\mathcal{V}$, it initially divides the image $\mathbf{I}$ into $N$ non-overlapping patches and projects them into the patch embeddings $\mathbf{P}_v\in \mathbb{R}^{N\times d_{v}}$. Concurrently, these embeddings $\mathbf{P}_v$, along with a learnable class token $\mathbf{P}_c$, are fed into $K$ transformer blocks (e.g., $K=12$ for ViT-B), processed in sequence. With ResNet as $\mathcal{V}$, $\mathbf{I}$ first goes through the convolution layers, followed by a multi-head attention pooling at the network's end, integrating a learnable class token $\mathbf{P}_c$ with patch (visual) features. Therefore, the image encoding process in CLIP can be illustrated as shown in Eq.~(\ref{visual_e}).
\begin{equation}
\label{visual_e}    
[\mathbf{X}_c, \mathbf{X}_v]=\mathcal{V}(\mathbf{I}).
\end{equation}

In line with this design, regardless of the selected visual encoder backbone, two types of tokens are finally generated: a class token $\mathbf{X}_c$ and visual tokens $\mathbf{X}_v$.

\textbf{Encoding Text:} The CLIP text encoder aims to generate feature representations for textual descriptions $T$. Firstly, each token in a textual description is a unique numeric ID encoded by byte pair encoding (BPE) algorithm. Then, numeric IDs are mapped to 512D word embedding vectors, which are passed into text encoder $\mathcal{L}$ stacking transformer layers. Note each text description is encompassed with the [SOS] and [EOS] tokens before token embedding. Finally, the features at the [EOS]
token position are as global sentence representation, followed by a linear project ($Proj_{txt}$):
\begin{align}  
& \mathbf{T}=\mathcal{L}(T), \\
& \mathbf{T}_c=Proj_{txt}(\mathbf{T}_{idx([EOS])}).
\end{align}
where $\mathbf{T}_{idx([EOS])}$ denots the index of [EOS] token and $\mathbf{T}_c$ is the global sentence representation.

\textbf{Zero-shot Classification:} The purpose of CLIP is pretrained on image-text pairs to predict whether an image matches a textual description. After pretraining, CLIP model can perform zero-shot recognition setting by comparing global image features $\mathbf{X}_c$ with global text features $\mathbf{T}_c$ encoded by image encoder $\mathcal{V}$ and text encoder $\mathcal{L}$ respectively. In zero-shot recognition, text prompts are manually created (i.e., "a photo of a \{cls\}") using class token, treated as $K$ weight vectors ($\mathbf{T}_c^1, ..., \mathbf{T}_c^K$), where the class token \{cls\} is replaced by the $K$ specific class name in corresponding prompt input, such as “dog”, “cat” or “car”. The prediction probability for the image $\textbf{I}$, which achieves the highest cosine similarity score ($\mathrm{sim}(\cdot,\cdot)$), is computed using a temperature factor $\tau$:
\begin{equation}
p({y=k}|\textbf{x})=\frac{exp(sim(\mathbf{X}_c,\mathbf{T}_c^k)/\tau )}{\textstyle\sum_{i=1}^{K}exp(sim(\mathbf{X}_c,\mathbf{T}_c^i))}.
\end{equation}


\subsection{Overall Structure of PM$^2$}
\label{s2.1}
In this section, we detail the process of the multi-modal model PM$^2$ for few-shot medical image classification, illustrated in Fig.~\ref{fig1}. As CLIP is used as backbone, there are two encoders in our PM$^2$, i.e., image encoder $\mathcal{V}$ and text encoder $\mathcal{L}$. Initially, a medical image is used as input for the visual component, and its associated category description serves as input for the text component. We denote the visual input as ${\mathbf{x}}_{\mathcal{V}}$ and the textual input as ${\mathbf{x}}_{\mathcal{L}}$, where each ${\mathbf{x}}_{\mathcal{V}}$ is paired with a corresponding text description ${\mathbf{x}}_{\mathcal{L}}$. These inputs undergo processing by their respective encoders to extract features. As depicted in Sec.~\ref{s2.0}, for both encoders, a single class token $\mathbf{X}_c$ and visual tokens $\mathbf{X}_v$ are obtained by image encoder $\mathcal{V}$, and $\mathbf{T}_c$, which is the global sentence representation, are obtained by text encoder $\mathcal{L}$.

\begin{figure}[t]
    \centering
    \includegraphics[scale=0.78]{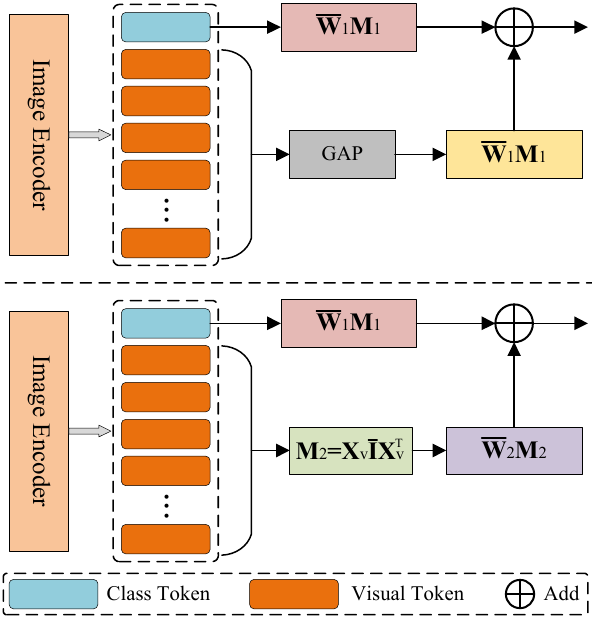}
    \caption{Schematic diagram of the visual classification head. The visual classification head aggregates the visual tokens output by the image encoder, differing from the naive use of global average pooling (GAP), as shown in the upper part of the diagram. This paper employs covariance pooling for second-order statistical modeling of visual tokens and combines this with the first-order information represented by a single class token for the visual component's prediction, as shown in the lower part of the diagram.}
    \label{fig2}
\end{figure}

For classification, we build a visual classification head to achieve a linear classifier on feature distribution, taking as input $\mathbf{X}_c$ and $\mathbf{X}_v$. In the top of image encoder, the head mainly captures the second-order statistics (i.e., covariance) inherent in visual tokens and make prediction on them, which are ignored in previous works. Besides, the head also considers the class token which is of a global image representation and the fist-order statistics (class token itself) of image. Thus such visual head is achieved by taking int consideration richer first- and second-order moments of features simultaneously. More details of visual classification head is provided in Sec.~\ref{s2.3} and Fig.~\ref{fig2}.

For individual text prediction, in the training stage, we can share or create its labels equal to corresponding image labels. However, in the inference stage, such scheme is prohibitive since image labels are unavailable as well in practice. As text prompts (Tab.~\ref{tab1} and Sec.~\ref{s2.2}) are treated as auxiliary samples to bolster the interplay between text and visual elements, we achieve this collaboration by using a shared linear classification head (shared parameters) for both class token $\mathbf{X}_c$ from image encoder and [EOS] token $\mathbf{T}_c$ from text encoder for individual predictions, referring to the top of Fig.~\ref{fig1}. By doing so, in the inference stage, predictions are exclusively made for medical images as text information has been absorbed into the shared linear classification head during training stage, simplifying the inference pipeline as well.

\subsection{Text Prompt}
\label{s2.2}
Creating effective prompts is essential for the model to generalize effectively with medical data. In this study, we explore five distinct prompt methods, building on existing research. These methods are classname, vanilla~\cite{radford2021learning}, hand-crafted~\cite{zhang2021tip}, class-specific descriptions generated via GPT~\cite{zhang2023prompt}, and the learnable CoOp approach~\cite{zhou2022learning}. Detailed descriptions of these methods, exemplified using the BACH dataset, are provided in Tab.~\ref{tab1}. The specifics for the first three methods are outlined in rows 2-4 of the table. Then, we elaborate on generating prompts with GPT and the learnable CoOp method.

\textbf{GPT Generates Prompt:} GPT~\cite{brown2020language} utilizes the transformer architecture, extensively pre-trained on vast text datasets for coherent text generation and understanding, making it valuable in various NLP tasks. In our research, we used the interactive tool ChatGPT~\footnote{https://chat.openai.com/}, based on GPT 3.5, to create prompts for different categories. We adopte two approaches to generating these prompts. Firstly, we confirm ChatGPT's grasp of our specific dataset before requesting prompts for the categories within it, as illustrated in the fifth row of our table. Secondly, we formulate prompts using a structured template, akin to the method used in CaFo~\cite{zhang2023prompt}, detailed in the sixth row of the table. For instance, the template “A caption of an image of a breast tissue with \{cls\}.” was employed.

\textbf{CoOp's Learnable Prompt:} The CoOp method~\cite{zhou2022learning} customizes CLIP-like V-L models for image recognition tasks. It does so by using learnable vectors as placeholders for context words in prompts, innovatively solving the problem of extensive prompt engineering time. In our study, following CoOp's framework, we design the inputs for the text encoder $\mathcal{L}$ as follows:
\begin{equation}
\label{e4}
\centering
    x_{\mathcal{L}}=[V]_{1}[V]_{2}...[V]_{M}[cls].
\end{equation}
In this setup, each $[V]_{m}$, where $m \in \{1,2,...,M\}$, matches the dimensionality of the word embedding (namely, 1024 or 512 for CLIP) and is shared among all [cls]. $M$ represents a hyperparameter that determines the count of context tokens. Our methodology draws upon the findings from the CoOp evaluation, employing a consistent contextual approach throughout.

\subsection{Visual Classification Head}
\label{s2.3}
In this section, we detail the visual classification head used for visual features extracted from vision encoder. Its structure is shown in Fig.~\ref{fig2}. In CLIP model, there are a single class token and multiple visual tokens. Given a class token $\mathbf{X}_c \in \mathbb{R}^d$ and a set of $N$ $d$-dimensional visual tokens $\mathbf{X}_v \in \mathbb{R}^{d\times N}$ output from the block right before classifier of pre-trained CLIP models. As mentioned above, linear probing usually takes as input class token $\mathbf{X}_c$ or average (mean) pooling of visual tokens $\mathbf{X}_v$ and generates prediction $\mathbf{y}_{pred}$ by
\begin{equation}\label{LP}
   \mathbf{y}_{pred} =  \mathbf{W}\phi(\mathbf{X}),
\end{equation}
where $\mathbf{W}\in \mathbb{R}^{C\times S}$ is the classifier weight and $C$ and $S$ are the number of classes and the dimension of representation $\phi(\mathbf{X})$. The $\phi(\mathbf{X})$ is equal to $\mathbf{X}_c$ or average pooling operation on $\mathbf{X}_v$. In Eq.~(\ref{LP}), the prediction $\mathbf{y}_{pred}$ linear probing makes deeply depends on representation $\phi(\mathbf{X})$. However, both the class token $\mathbf{X}_c$ itself and average pooling of visual tokens $\mathbf{X}_v$ are first-order feature statistics which is a simple mean point (the upper part of Fig.~\ref{fig2}), limiting its modeling ability.
Therefore, for making the prediction more precise, we model feature distribution on $\mathbf{X}_v$ to generate stronger representation $\theta(\mathbf{X}_v)$ which characterizes the full picture of features. Based on $\theta(\mathbf{X}_v)$, we obtain
\begin{equation}\label{dist}
   \mathbf{y}_{pred} =  \mathbf{\overline{W}}\theta(\mathbf{X}_v),
\end{equation}
In practice, the function $\theta(\mathbf{X}_v)$ is unknown usually. According to the classical probability theory, $\theta(\mathbf{X}_v)$ can be defined by its characteristic function $\varphi_\textbf{X}(t)$, computed by
\begin{equation}\label{characteristic}
    \begin{aligned}
        \varphi_{\mathbf{X}_v}(t) = E\left[\sum_{k=0}^{\infty}\frac{(it\mathbf{X}_v)^k}{k!}\right] = \sum_{k=0}^{\infty}\frac{(it)^k}{k!}E\left[\mathbf{X}_v^k\right] \\
   =1+(it)E\left[\mathbf{X}_v\right]+\frac{(it)^2}{2!}E\left[\mathbf{X}_v^2\right]+\cdot\cdot\cdot  ,
    \end{aligned}
\end{equation}
where $i$ and $t$ are the imaginary unit and argument
of the characteristic function respectively. $E\left[\mathbf{X}_v^k\right]$ refers to the $k^{th}$-order moment of $\mathbf{X}_v$. For simplicity, $\varphi_{\textbf{X}_v}(t)$ is rewritten as
\begin{equation}\label{characteristic_V2}
    \begin{aligned}
        \varphi_{\mathbf{X}_v}(t) 
            =1+\omega_1\mathbf{M}_1+\omega_2\mathbf{M}_2+ \cdot\cdot\cdot  =1+\sum_{k=1}^{\infty}\omega_k\mathbf{M}_k,
    \end{aligned}
\end{equation}
where $\omega_k$ is the coefficient of moment of order $k$ (i.e., $E\left[\mathbf{X}_v^k\right]$) and $\mathbf{M}_k=E\left[\mathbf{X}_v^k\right]$.

So far, in Eq.~(\ref{dist}), feature distribution $\theta(\mathbf{X}_v)$ can be represented by characteristic function $\varphi_{\textbf{X}_v}(t)$ in Eq.~(\ref{characteristic_V2}), i.e.,
\begin{equation}\label{LP_V2}
   \mathbf{y}_{pred} =  \mathbf{\overline{W}}\left(\sum_{k=1}^{\infty}\omega_k\mathbf{M}_k\right) = \sum_{k=1}^{\infty}\left(\mathbf{\overline{W}}_k\mathbf{M}_k\right).
\end{equation}
In Eq.~(\ref{LP_V2}), we omit the constant term. It can be seen that, based on Eq.~(\ref{LP_V2}), linear probing makes prediction based on different moments of order which better characterize feature distribution more reasonably than average pooling of $\mathbf{X}_v$ (mean point). As such, Eq.~(\ref{LP_V2}) is able to predict more accurately.
In our case, we only consider the first- and second-order moments ($\mathbf{M}_1$ and $\mathbf{M}_2$) of $\mathbf{X}_v$ for efficiency again. Then finally prediction is achieved by
\begin{equation}\label{LP_V3}
   \mathbf{y}_{pred} = \sum_{k=1}^{2}\left(\mathbf{\overline{W}}_k\mathbf{M}_k\right)=\mathbf{\overline{W}}_1\mathbf{M}_1 + \mathbf{\overline{W}}_2\mathbf{M}_2.
\end{equation}
The $\mathbf{\overline{W}}_1$ and $\mathbf{\overline{W}}_2$ are associated classifier parameters. As class token $\mathbf{X}_c$ can represent holistic image, for making better use of it,
$\mathbf{M}_1$ in Eq.~(\ref{LP_V3}) is replaced with it ($\mathbf{M}_1=\mathbf{X}_c$, the lower part of Fig.~\ref{fig2}).
The $\mathbf{M}_2$ is computed by covariance pooling:
\begin{equation}
\label{e6}
\centering   
\mathbf{M}_2 = \mathbf{\Sigma}=\mathbf{X}_v\mathbf{\bar{I}} \mathbf{X}_v^{T}.
\end{equation}
where $\mathbf{\bar{I}}=\frac{1}{N} (\mathbf{I}-\frac{1}{N} \mathbf{1} )$, $\mathbf{I}$ and $\mathbf{1}$ are the $N \times N$ identity matrix and the all-ones matrix, respectively.


Additionally, following the suggestion of~\cite{li2018towards}, we further normalize the $\mathbf{\Sigma}$ ($\mathbf{M}_2$) using matrix square root to enhance its modeling ability, which actually performs element-wise square root on its eigenvalues. As such, we need to compute the eigenvalues of the covariance matrix $\mathbf{\Sigma}$. However, exact computation of eigenvalues depends deeply on EIG or SVD, which is limitedly supported on GPU parallel processing. Alternatively, a fast, approximate matrix square root normalization, called Newton-Schulz iteration, is employed to achieve the same purpose.
The Newton-Schulz iteration only involves matrix product, naturally avoiding the GPU-unfriendly calculation, thus helpful for realizing parallel computing efficiently. Due to Newton-Schulz iteration only converges locally, covariance matrix $\mathbf{\Sigma}$ needs to be pre-normalized by its trace:
\begin{equation}
    \mathbf{Q}=\frac{1}{\mathrm{tr}(\mathbf{\Sigma})}\mathbf{\Sigma}.
\end{equation}

Then, given $\mathbf{J}_{0}=\mathbf{Q}$ and $\mathbf{P}_{0}=\mathbf{I}$, the Newton-Schulz iteration estimates the square root
$\mathbf{J}$ of $\mathbf{Q}$ by the following form:
\begin{align}
\begin{split}
\mathbf{J}_{k} = \frac{1}{2}\mathbf{J}_{k-1}(3\mathbf{I} - \mathbf{P}_{k-1}\mathbf{J}_{k-1}), \\
\mathbf{P}_{k} = \frac{1}{2}(3\mathbf{I} - \mathbf{P}_{k-1}\mathbf{J}_{k-1})\mathbf{P}_{k-1}.
\label{iSQRT}
\end{split}
\end{align}

After a small number of iterations ($k=3$), $\mathbf{J}_{k}$ is converged approximately to $\mathbf{\Sigma}^{\frac{1}{2}}$. As pre-normalization is introduced before iterations, as a result, post-compensation is also needed to counteract its adverse
effect, preducing square root normalized covariance matrix:
\begin{equation}
    \mathbf{\Sigma}^{\frac{1}{2}}=\sqrt{\mathrm{tr}(\mathbf{\Sigma})}\mathbf{J}_{k}.
\end{equation}

Obviously, $\mathbf{\Sigma}^{\frac{1}{2}}$ is a symmetric matrix. Therefore, we concatenate its upper triangular entries to form the final feature representation of the patch tokens, which are submitted to the subsequent classifier (a fully-connected layer followed by softmax). And $\mathbf{M}_2$ in Eq.~(\ref{LP_V3}) is replaced with $\mathbf{\Sigma}^{\frac{1}{2}}$ to predict $\mathbf{y}_{pred}$.

\section{Experiments}
This section details the outcomes of our method via various experiments. We first describe the dataset used in our research in Sec.~\ref{3.1} and then outline the specific implementation details in Sec.~\ref{3.2}. Following this, we conduct ablation studies to examine the effects of different components, such as text prompts and visual classification heads. In the final part, we apply contemporary multi-modal techniques to medical image datasets and compare these with the methodologies discussed in this paper.
\subsection{Datasets}
\label{3.1}
The BACH (Breast Cancer Histology) dataset~\cite{aresta2019bach} is a collection of histopathological images derived from breast cancer biopsies, aiming primarily at classifying various breast cancer subtypes through identifiable visual patterns in the images. It is carefully organized into four categories: Normal, Benign, In Situ Carcinoma, and Invasive Carcinoma, with each class containing 100 meticulously selected histopathological images.

The Figshare MRI brain tumor dataset~\cite{2015Correction} consists of 3064 T1-weighted contrast-enhanced images from 233 patients, divided into three brain tumor categories: meningioma, glioma, and pituitary tumors, with respective image counts of 708, 1426, and 930. This dataset is available in MATLAB data format (*.mat file), with each image measuring 512 × 512 pixels and carefully annotated by expert radiologists.

The Diabetic Retinopathy (DR) dataset~\cite{dekhil2019deep}, meticulously captured at Aravind Eye Hospital, is comprehensively analyzed and classified by medical experts according to the severity of lesions. This dataset encompasses 3662 high-resolution color retinal images, systematically sorted into five categories: non-proliferative DR, mild non-proliferative DR, moderate non-proliferative DR, severe non-proliferative DR, and proliferative DR. It comprises 1805 images of non-proliferative DR, 370 of mild non-proliferative DR, 999 of moderate non-proliferative DR, and 193 combining severe non-proliferative with proliferative DR. Furthermore, there are 295 images categorized as general DR.

In utilization of these datasets, we partition them into training and validation sets with a division ratio of 7:3.

\begin{table*}[H]
\centering
\caption{Evaluate different text prompts with ResNet as the backbone.}\label{tab2}
\begin{tabular}{ccccccc}
\hline
\multicolumn{7}{c}{Backbone is ResNet50} \\ \hline
modal     &  text prompt   & 1-shot & 2-shot &  4-shot & 8-shot & 16-shot \\ \hline
uni modal  & none & 38.33 &  42.78  &  50.55  &  53.89  & 65.00    \\
\multirow{7}{*}{multi modal} & classname    &  38.33  &  43.33  &  51.11  & 54.17  & 66.39   \\
 &  vanilla    & 39.17  &  44.17  & 51.67&  54.16  &  65.28  \\
 &  hand-crafted &  41.11  & 47.22  & 51.67  & 55.56  &  \underline{66.95}   \\
 &  GPT Prompt 0       &  41.66  & \underline{45.28}  &  51.67  & 56.39  &  \textbf{67.22}   \\
 & GPT Prompt 1         &  41.39  & 44.44  &  50.00     & 56.39  & 66.94   \\
 &  CoOp$_{4}$     &  \textbf{44.45}  &  \textbf{48.61}  & \textbf{52.78}  & \underline{56.67}  &  65.83  \\ 
 & CoOp$_{16}$   & \underline{42.50}   & 45.00   &  \underline{51.94}  &  \textbf{57.50}   &  66.39   \\ \hline \hline
\multicolumn{7}{c}{Backbone is ResNet101} \\ \hline
uni modal   & none          & 40.56 & 50.83 &  50.00  &  58.34  & 66.11 \\
\multirow{7}{*}{multi modal} & classname     & 40.83 & 50.83 &  50.28  &  58.89  & 66.39   \\
 &  vanilla      & 42.22 & 50.66 &  51.39  &  58.06  & 66.95  \\
 &  hand-crafted & 41.67 & 51.39 &  52.50  &  \textbf{62.45}  & 67.22   \\
 &  GPT Prompt 0        & 42.78 & 49.72 &  51.94  &  58.89  & 67.22   \\
 & GPT Prompt 1         & 39.16 & 51.11 &  50.83  &  58.33  & 66.11   \\
 &  CoOp$_{4}$   & \underline{44.72} & \textbf{53.61} &  \underline{53.89}  &  60.00  & \textbf{68.61}   \\
 & CoOp$_{16}$   & \textbf{49.72} & \underline{53.33} &  \textbf{55.28}  &  \underline{60.84}  & \underline{67.78}  \\ \hline
\end{tabular}
\end{table*}

\begin{table*}[H]
\centering
\caption{Evaluate different text prompts with ViT-B as the backbone.}\label{tab4}
\begin{tabular}{ccccccc}
\hline
\multicolumn{7}{c}{Backbone is ViT-B/32} \\ \hline
modal     &  text prompt   & 1-shot & 2-shot &  4-shot & 8-shot & 16-shot \\ \hline
uni modal   & none          & 43.61 & 45.28 &  50.55  &  54.44  & 64.17 \\
\multirow{7}{*}{multi modal} & classname     & 43.33 & \underline{46.94} &  52.50  &  54.44  & 65.83   \\
 &  vanilla      & 42.50 & 44.45 &  53.06  &  54.17  & 65.00  \\
 &  hand-crafted & \underline{45.28} & 46.67 &  \textbf{54.44}  &  \underline{56.67}  & 65.83   \\
 &  GPT Prompt 0        & 44.44 & 45.55 &  53.61  &  56.11  & 66.38   \\
 & GPT Prompt 1         & 40.28 & 44.16 &  51.67  &  52.50  & 63.89   \\
 &  CoOp$_{4}$   & \textbf{47.50} & \textbf{49.17} &  52.50  &  55.83  & \textbf{66.67}  \\
 & CoOp$_{16}$   & 45.00 & 45.83 &  \underline{53.89}  &  \textbf{57.78}  & \underline{66.39}   \\ \hline \hline
\multicolumn{7}{c}{Backbone is ViT-B/16} \\ \hline
uni modal   & none          & 38.34 & 44.17 &  48.33  &  51.39  & 67.50 \\
\multirow{7}{*}{multi modal} & classname     & 38.61 & 45.00 &  50.00  &  52.22  & 67.78   \\
 &  vanilla      & 38.89 & 44.45 &  49.17  &  51.94  & 67.78  \\
 &  hand-crafted & 41.67 & 46.39 &  51.39  &  55.28  & 69.17   \\
 &  GPT Prompt 0        & 38.06 & 45.83 &  \textbf{53.33}  &  54.72  & \underline{70.00}   \\
 & GPT Prompt 1         & \textbf{45.83} & 46.94 &  51.94  &  53.89  & 69.16   \\
 &  CoOp$_{4}$   & \underline{44.17} & \textbf{51.94} &  52.50  &  \underline{56.39}  & 69.44   \\
 & CoOp$_{16}$   & 41.67 & \underline{48.89} &  \underline{52.78}  &  \textbf{56.67}  & \textbf{70.00}  \\ \hline
\end{tabular}
\end{table*}

\subsection{Implementation Details}
\label{3.2}
We utilize pre-trained CLIP~\cite{radford2021learning} models that contain dual encoders for encoding medical images and text, significantly streamlining the training process. For efficiency, the CLIP models are frozen during training, and only the prompt if involved (i.e., $[V]_{m}$ in CoOP) and classifiers are updated. Importantly, in evaluating text prompts, a shared fully connected layer serves as the classifier for individual [EOS] token and class token. For optimizing the network, the AdamW~\cite{kingma2014adam} algorithm is employed, combined with a warmup phase and followed by a cosine annealing learning rate schedule. The warmup period covers 50 iterations, with the entire training spanning 12,800 iterations. We train our model with a total batch size of 2 and conduct a systematic search to optimize learning rate and weight decay, exploring ranges of [0.001, 0.0001] for learning rate and [0.0, 0.01, 0.0001] for weight decay. For the visual input, images are cropped and resized to 224$\times$224.

Additionally, it is important to highlight our rigorous methodology in using three different random seeds to sample a specified number of few-shot instances \{1, 2, 4, 8, 16\} from the training set, as outlined in the methodologies of references~\cite{radford2021learning,zhou2022learning,lin2023multimodality}. The results presented in this study are averaged across three separate runs.

\subsection{Effect of Text Prompts}

In this section, we assess the effectiveness of text prompts
on the BACH dataset. In multi-modal settings, various text prompts under ResNet (ResNet50, ResNet101) and ViT (ViT-B/32, ViT-B/16) architectures are evaluated on different shots of \{1, 2, 4, 8, 16\}. By the way, neither single-modal settings nor multi-modal settings make predicts on second-order feature distribution. The only difference between them is whether text prompts are used. The corresponding ablation results are listed in Tabs.~\ref{tab2} and~\ref{tab4}, respectively.

In the context of both tables, `classname' means only the class name used as text input. Prompt types about vanilla, hand-crafted, GPT prompts 0 and GPT prompts 1 can be found in Tab.~\ref{tab1}. We also implement the learnable CoOp method~\cite{zhou2022learning}, with $M$ learnable context vectors tailored to CoOp's specifications. Here, ablation experiments are conducted with two configurations of $M=\{4,16\}$ context vectors (CoOP$_4$ and CoOP$_{16}$ in the second column). In these tables, bold typeface indicates the best performance, while underlined typeface indicates the second-best. Additionally, `uni modal' refers to using only the visual mode, and `multi modal' indicates the use of both text and visual modes.

Tab.~\ref{tab2} showcases the effectiveness of different text prompting methods using ResNet50 and ResNet101 as visual backbones. Among all text prompts, the CoOp~\cite{zhou2022learning} method emerges as a consistent top performer, especially in higher shot settings like 8-shot and 16-shot. For example, with ResNet50 as the backbone, CoOp settings always yield the best and second-best results. While the 8-shot results with ResNet101 diverge from this pattern, CoOp settings still dominate the optimal and near-optimal outcomes, a testament to CoOp's adaptability in handling sampled data to improve model generalization. Additionally, the hand-crafted and GPT prompt methods show strong performance in particular cases, such as GPT prompt 0 and the hand-crafted method respectively record 67.22\% and 66.95\% in top and second-best results under ResNet50 as the backbone. This highlights the value of well-crafted text prompts and the use of pre-trained language models in boosting model recognition accuracy.

The results using ViT-B/32 and ViT-B/16 as visual backbones are detailed in Tab.~\ref{tab4}. The CoOp~\cite{zhou2022learning} method continues to excel as a leading performer in most cases. With ViT-B/32, CoOp achieves six top and near-top results across various shot settings. In scenarios where ViT-B/16 is the backbone, CoOp secures seven leading and near-leading outcomes, highlighting its strong performance in medical image classification. Notably, the GPT prompt and hand-crafted methods consistently rank as the best or second-best, reflecting the crucial role of expertly crafted prompts in driving superior results.

The analysis of results from the tables clearly demonstrates that incorporating textual information significantly and consistently enhances performance across all prompt methods compared to a single-modal approach in terms of all shot settings and network types. 
The CoOp method distinguishes itself by consistently outperforming others, owing to its learnable prompts that adapt better to specific visual models and training datasets. This adaptability is particularly beneficial for diverse task types and few-shot training contexts. Similarly, improvements with the hand-crafted and GPT prompt methods underscore the effectiveness of well-designed text prompts and those created using sophisticated language models, especially in certain visual backbone and shot settings. These results underscore the vital role of textual prompts in enhancing medical image recognition performance on such challenging few-shot settings with limited samples.

\subsection{Effect of Second-order Statistics}
This section focuses on evaluating feature distribution related to the visual classification head shown in Fig~\ref{fig2}. We utilize ViT-B/16 as backbone, coupled with the CoOp text prompt method where fewer $M=4$ context vectors are introduced. The dimension of visual tokens is reduced to 96 via a $1\times 1$ convolution layer for efficiency.
We first explore the impact of second-order pooling, as detailed in Tab.~\ref{tab6}. The 'cls' in Tab.~\ref{tab6} represents only a single class token is utilized for prediction in our visual classification head. Using visual tokens with only first-order aggregation via global average pooling shows lower performance than class token method, especially in limited shot scenarios of \{1, 2\}. An exception is observed in larger shot settings (e.g., 16-shots), where visual tokens surpasses class token by 2.5\%. This suggests ViT's potential limitation in the reasonable use of visual tokens under a downstream task, a gap not fully addressed by limited samples and simple aggregation of such tokens. Further, visual tokens are aggregated by the second-order pooling not naive average pooling, forming our final visual classification head. It results in a 74.72\% accuracy in the 16-shot setup, a 5.28\% improvement over class token-only methods. Moreover, 'cls+visual\_so' shows a 7.78\% and 8.33\% gains over baseline ('cls') in the 4-shot and 8-shot settings, respectively, and yields improvements of 2.77\% and 3.9\% in the 1-shot and 2-shot settings. These results demonstrate the effectiveness of the second-order pooling approach in few-shot scenarios, extracting valuable statistical information inherent in visual tokens and significantly boosting performance.

\begin{table}[t]
	\centering
 \caption{Evaluate the impact of second-order pooling on performance.}
 \label{tab6}
\begin{tabular}{cccccc} \hline
 Method & 1-shot & 2-shot & 4-shot & 8-shot & 16-shot \\ \hline
cls                           & 44.17  & 51.94  & 52.50  & 56.39  & 69.44   \\
cls+visual\_avg               & 39.17  & 49.17  & 52.50  & 56.11  & 71.94    \\
cls+visual\_so                & \textbf{46.94}  & \textbf{55.84}  & \textbf{60.28} & \textbf{64.72}  & \textbf{74.72}  \\ \hline
\end{tabular}
\end{table}

The influence of dimensions of second-order pooling on performance is significant. Extremely high dimensions can lead to overfitting and bring much computation, while very low dimensions might cause a loss of information. In light of this, we make an ablation study to evaluate its effects, whose setting and results are in Tab.~\ref{tab7}. The dimension of visual tokens $\mathbf{X}_c$ outputted from ViT-B/16 is of 768, which is very high for computing a covariance matrix with size of  $768\times 768$. Thus, we have to reduce its dimension before second-order pooling.
Tab.~\ref{tab7} showcases how different dimensional settings yield varying results. For example, setting the dimension to 48 results in a peak performance of 48.61\% in the 1-shot scenario, while a dimension of 64 leads to the best performance of 76.11\% in the 16-shot scenario. Remarkably in addition to too high dimensions bringing overfitting challenges, especially with smaller sample sizes, we also face the covariance estimate of not robust to high dimensional problems with a small number of samples~\cite{won2013condition,wang2016raid}. As illustrated, dimensions set at higher 384 and 256 result in 1-shot performances of 37.50\% and 37.78\%, respectively, which are even lower than those in the single-modal approach. By comparison and analysis, dimension of 96 is selected as final setting for the computation of second-order pooling in visual classification head.

\begin{table}[t]
	\centering
 \caption{Comparison of second-order pooling performance under different dimensions.}
 \label{tab7}
\begin{tabular}{cccccc} \hline
 Dim. & 1-shot & 2-shot & 4-shot & 8-shot & 16-shot \\ \hline
384               & 37.50  & 46.39  & 55.00  & \textbf{65.28}  & 75.00   \\
256               & 37.78  & 47.22  & 56.94  & 64.45  & 75.56    \\
128               & 45.00  & 54.72  & 59.72  & 65.00  & 74.44       \\
96                & 46.94  & \textbf{55.84}  & \textbf{60.28} & 64.72  & 74.72 \\ 
64                & 47.5   & 55.00 & 57.78  & 64.72  & \textbf{76.11}  \\ 
48                & \textbf{48.61}  & 55.83 & 58.33  & 64.17  & 75.28  \\ 
\hline
\end{tabular}
\end{table}

\subsection{Comparison with Other Multi-modal Methods}

The comparisons include independent implementations of both Tip-Adapter (abbreviated as Tip)~\cite{zhang2021tip} and its refined version, Tip-Adapter-F (abbreviated as Tip-F)~\cite{zhang2021tip}, as well as CoOp~\cite{zhou2022learning} and its enhanced version, CoCoOp~\cite{zhou2022conditional}. These methods are rigorously tested across three diverse datasets, allowing for a thorough comparison with our proposed PM$^2$. For comparative analysis, CoOp is consistently used as the text prompt method, with the learnable word vector length $M$ set to 4 as well. The ViT-B/16 is also utilized as the visual backbone. Note that although the dimension of visual tokens $\mathbf{X}_c$ set to 96 reports the best performance, for further efficiency, we adopt dimension of 64 instead of 96 to compute second-order pooling for comparisons with state-of-the-art methods. So our PM$^2$ can potentially improve performance when dimension is set to 96.


Tab.~\ref{CBACH} offers a comparative analysis of various methods applied to the BACH dataset, showcasing the classification accuracy of different multi-modal approaches across diverse shot scenarios. Our method stands out significantly, topping the results in nearly all sample size, except in the 1-shot setting where it falls short by just 2.5\% compared to the CoCoOp\cite{zhou2022conditional} method. But the performance of our approach is especially noteworthy in the 8-shot and 16-shot settings, where it achieves respectively impressive accuracy rates of 64.72\% and 76.11\%, significantly outperforming CoCoOp by a large margin of 12.79\% and 25.28\%. Compared to Tip, Tip-F and CoOP, our advantages are also obvious in all shot settings.

\begin{table}[t]
	\centering
 \caption{Comparison with other multi-modal methods on BACH dataset.}
 \label{CBACH}
\begin{tabular}{cccccc} \hline
Method & 1-shot  & 2-shot    & 4-shot  & 8-shot    & 16-shot    \\ \hline
\multicolumn{6}{c}{CLIP Zero shot~\cite{radford2021learning}: 20.00}   \\ \hline
Tip~\cite{zhang2021tip}    & 42.50 & 44.17 & 49.17 & 49.17 & 52.50 \\
Tip-F~\cite{zhang2021tip}  & 35.83 & 54.17 & 46.67 & 55.83 & 55.83 \\
CoOp~\cite{zhou2022learning}           & 31.10 & 36.10 & 34.73 & 49.43 & 55.60 \\
CoCoOp~\cite{zhou2022conditional}         & \textbf{50.00} & 51.93 & 54.43 & 51.93 & 50.83 \\ \hline
PM$^2$          & 47.5   & \textbf{55.00} & \textbf{57.78}  & \textbf{64.72}  & \textbf{76.11} \\ \hline
\end{tabular}
\end{table}

Tab.~\ref{Cbrain} illustrates the comparative analysis of various methods on the brain tumor dataset, where our method consistently exceeds the performance of others in every shot setting. This is particularly apparent in the 8-shot and 16-shot scenarios, where our method achieves impressive gains, outperforming all competing approaches. In a similar vein, Tab.~\ref{Cdr} showcases our method's results on the DR dataset. Here, our PM$^2$ is highly competitive, lagging behind the Tip method by only 0.57\% in the 4-shot setting, yet surpassing all other methods in the remaining shot settings. 

\begin{table}[t]
	\centering
 \caption{Comparison with other multi-modal methods on brain tumor dataset.}
 \label{Cbrain}
\begin{tabular}{cccccc} \hline
Method & 1-shot  & 2-shot    & 4-shot  & 8-shot    & 16-shot    \\ \hline
\multicolumn{6}{c}{CLIP Zero shot~\cite{radford2021learning}: 41.36}   \\ \hline
Tip~\cite{zhang2021tip}       & 45.44 & 52.11 & 49.39 & 51.43 & 72.38 \\
Tip-F~\cite{zhang2021tip}             & 44.49 & 43.54 & 52.93 & 59.59 & 65.58 \\
CoOp~\cite{zhou2022learning}          & 40.03 & 43.23 & 53.27 & 61.60 & 60.00 \\
CoCoOp~\cite{zhou2022conditional}     & 34.27 & 31.43 & 40.97 & 31.27 & 47.70 \\ \hline
PM$^2$                                  &  \textbf{54.69} & \textbf{55.28} & \textbf{64.99} & \textbf{78.87} & \textbf{84.58} \\
\hline
\end{tabular}
\end{table}

\begin{table}[t]
	\centering
 \caption{Comparison with other multi-modal methods on DR dataset.}
 \label{Cdr}
\begin{tabular}{cccccc} \hline
Method & 1-shot  & 2-shot    & 4-shot  & 8-shot    & 16-shot    \\ \hline
\multicolumn{6}{c}{CLIP Zero shot~\cite{radford2021learning}: 49.45}   \\ \hline
Tip~\cite{zhang2021tip}            &  62.32 & 56.48 & \textbf{69.34} & 62.23 & 66.61 \\
Tip-F~\cite{zhang2021tip}          &  60.95 & 58.21 & 67.15 & 66.97 & 67.43 \\
CoOp~\cite{zhou2022learning}       &  53.60 & 56.27 & 58.90 & 60.03 & 63.33 \\
CoCoOp~\cite{zhou2022conditional}  &  50.23 & 49.32 & 50.12 & 58.07 & 55.70 \\ \hline
PM$^2$                               &  \textbf{66.58} & \textbf{64.38} & 68.77 & \textbf{68.28} & \textbf{72.17} \\
\hline
\end{tabular}
\end{table}

\section{Discussion}
Our study has shown the effectiveness of using multi-modal methods for few-shot classification tasks in medical imaging. Nevertheless, the full fine-tuning of pre-trained networks is problematic due to limitations inherent in existing datasets. The extensive number of parameters to be updated poses a risk of overfitting. For example, with the BACH dataset, 1-shot learning yields merely 4 training samples, rendering the thorough fine-tuning of backbone parameters across a wide range unfeasible in such contexts.

Despite various challenges in medical image classification, our method demonstrates superior performance, effectively using up to 16 samples per category. This exploration sets the stage for further progress in the field. Envision a future where datasets don’t need extensive annotations for pathological image recognition. With as few as 16, or possibly even fewer, meticulously chosen samples, we can develop highly precise auxiliary diagnostic models for specialized tasks. Moreover, the ability to learn from a limited number of samples significantly reduces the need for extensive computational resources. This simplification in model development not only facilitates easier creation of custom models by different organizations but also plays a crucial role in enhancing data security and minimizing the risk of data leakage.

In future research, our aim is to investigate the potential impact of multi-modal solutions in the field of medical images. This entails designing more effective prompt methods~\cite{jia2022visual,wang2023review,bar2022visual}, strengthening the interaction between visual and textual modalities~\cite{zhou2022conditional,zhang2023prompt}, and incorporating new modalities~\cite{lin2023multimodality,zhao2023bubogpt} to improve model ability to capture comprehensive and relevant information. Additionally, our approach can be viewed as joint learning for text and image classification tasks. We will also delve deeper into the potential advantages of multi-task joint learning for this specific project.

\section{Conclusion}
Computer-aided diagnosis is of paramount importance. However, existing methodologies often rely on the availability of large-scale annotated datasets, which can be impractical in real-world application scenarios. In response, we propose a novel multi-modal approach PM$^2$ for few-shot medical image classification, leveraging recent advancements in multi-modal learning. Our approach integrates textual labels as supplementary samples to enhance model performance, and we empirically investigate the performance impact of various prompt learning techniques. Furthermore, to fully exploit image information, we harness visual tokens within the visual component to enrich the feature space for classification. We combine the second-order features derived from these tokens with the first-order information represented by class token, thereby enhancing the model's representational capacity. Finally, we conduct comprehensive comparisons with other state-of-the-art multi-modal approaches across three distinct medical datasets. The experimental results underscore the effectiveness of our approach, as it consistently achieves state-of-the-art performance levels. 


\section*{CRediT authorship contribution statement}
Zhenwei Wang: Methodology, Validation, Writing, Data curation. Qiule Sun: Methodology, Validation, Writing. Bingbing Zhang: Conceptualization, Supervision, Methodology, Writing-review \& editing. Pengfei Wang: Conceptualization, Methodology, Writing-review \& editing. Jianxin Zhang: Conceptualization, Methodology, Writing-review \& editing. Qiang Zhang: Supervision, Writing-review \& editing.

\section*{Declaration of competing interest}
The authors declare that they have no known competing financial interests or personal relationships that could have appeared to influence the work reported in this paper.

\section*{Data availability}
The data that support the findings of this study are openly available online at: \href{https://iciar2018-challenge.grand-challenge.org/}{https://iciar2018-challenge.grand-challen
ge.org/}, \href{https://figshare.com/articles/dataset/brain\_tumor\_dataset/1512427}{https://figshare.com/articles/dataset/brain\_tumor\_da
taset/1512427} and \href{https://www.kaggle.com/c/aptos2019-blindness-detection/overview}{https://www.kaggle.com/c/aptos2019-blin
dness-detection/overview}.

\section*{Acknowledgements}
This work was supported in part by the National Key Research and Development Program of China (No.2021ZD011
2400), the Liaoning Revitalization Talents Program (No.XL
YC2008017), the National Natural Science Foundation of China (No.61972062), the Applied Basic Research Project of Liaoning Province (No.2023JH2/101300191), Science and Technology Development Program Project of Jilin Province (No.20230201111GX).

\bibliography{example}

\end{document}